\documentclass[]{IAC_style}
\usepackage{array}
\usepackage{tabu}
\begin{document}
\IACpaperyear{2024} 
\IACpapernumber{IAC-24-C2-3-4-x87263} 
\IAClocation{Milan, Italy} 
\IACdate{14-18 October 2024} 

\IACcopyrightB{Mr. Yuxing Wang}

\title{Morphology and Behavior Co-Optimization of Modular Satellites for Attitude Control}

\IACauthor{Yuxing Wang$^{\orcidlink{0000-0000-0000-0000}}$}{1}{0}
\IACauthor{Jie Li$^{\orcidlink{0000-0000-0000-0000}}$}{2}{0}
\IACauthor{Cong Yu$^{\orcidlink{0000-0000-0000-0000}}$}{3}{0}
\IACauthor{Xinyang Li$^{\orcidlink{0000-0000-0000-0000}}$}{4}{0}
\IACauthor{Simeng Huang$^{\orcidlink{0000-0000-0000-0000}}$}{5}{0}
\IACauthor{Yongzhe Chang$^{\orcidlink{0000-0000-0000-0000}}$}{6}{0}
\IACauthor{Xueqian Wang$^{\orcidlink{0000-0000-0000-0000}}$}{7}{0}
\IACauthor{Bin Liang$^{\orcidlink{0000-0000-0000-0000}}$}{8}{0}

\IACauthoraffiliation{AI \& Robot Laboratory, Shenzhen International Graduate School (SIGS), Tsinghua University, University Town of Shenzhen, Nanshan District, Shenzhen, P.R. China, 518055 \normalfont{E-mail:~\authormail{wangyuxing@sz.tsinghua.edu.cn}}}
\IACauthoraffiliation{Institute of Distributed Spacecraft Systems Technology, School of Aerospace Engineering, Beijing Institute of Technology, Haidian District, Beijing, P.R. China, 100084 \normalfont{E-mail:~\authormail{lijie bit@bit.edu.cn}}}
\IACauthoraffiliation{AI \& Robot Laboratory, Shenzhen International Graduate School (SIGS), Tsinghua University, University Town of Shenzhen, Nanshan District, Shenzhen, P.R. China, 518055 \normalfont{E-mail:~\authormail{yuc23@mails.tsinghua.edu.cn}}}
\IACauthoraffiliation{AI \& Robot Laboratory, Shenzhen International Graduate School (SIGS), Tsinghua University, University Town of Shenzhen, Nanshan District, Shenzhen, P.R. China, 518055 \normalfont{E-mail:~\authormail{xy-li23@mails.tsinghua.edu.cn}}}
\IACauthoraffiliation{Haidian District, Beijing, P.R. China, 100084 \normalfont{E-mail:~\authormail{simenghuang@outlook.com}}}
\IACauthoraffiliation{AI \& Robot Laboratory, Shenzhen International Graduate School (SIGS), Tsinghua University, University Town of Shenzhen, Nanshan District, Shenzhen, P.R. China, 518055 \normalfont{E-mail:~\authormail{changyongzhe@sz.tsinghua.edu.cn}}}
\IACauthoraffiliation{AI \& Robot Laboratory, Shenzhen International Graduate School (SIGS), Tsinghua University, University Town of Shenzhen, Nanshan District, Shenzhen, P.R. China, 518055 \normalfont{E-mail:~\authormail{wang.xq@sz.tsinghua.edu.cn}}}
\IACauthoraffiliation{Department of Automation, Tsinghua University, Haidian District, Beijing, P.R. China, 100084 \normalfont{E-mail:~\authormail{bliang@tsinghua.edu.cn}}}

\abstract{
The emergence of modular satellites marks a significant transformation in spacecraft engineering, introducing a new paradigm of flexibility, resilience, and scalability in space exploration endeavors. In addressing complex challenges such as attitude control, both the satellite's morphological architecture and the controller are crucial for optimizing performance. Despite substantial research on optimal control, there remains a significant gap in developing optimized and practical assembly strategies for modular satellites tailored to specific mission constraints. This research gap primarily arises from the inherently complex nature of co-optimizing design and control, a process known for its notorious bi-level optimization loop. Conventionally tackled through artificial evolution, this issue involves optimizing the morphology based on the fitness of individual controllers, which is sample-inefficient and computationally expensive. In this paper, we introduce a novel gradient-based approach to simultaneously optimize both morphology and control for modular satellites, enhancing their performance and efficiency in attitude control missions. Specifically, to address the 3D modular design space, we introduce a scalable 3D Neural Cellular Automata (NCA) as the design policy. This NCA encodes local update rules in a Multilayer Perceptron (MLP) neural network, generating different developmental outcomes using a smaller set of trainable parameters. This enables gradient-based optimization and facilitates design pattern regeneration. For control optimization, we utilize a reaction-wheel-based attitude control system modeled with a neural network that shares parameters with the design policy. This shared parameterization enhances the network's generalization ability across different configurations. We integrate design and control processes into a unified Markov Decision Process (MDP), which enables the exploration of a broad design space. To achieve concurrent training of both policies, we employ Twin Delayed Deep Deterministic (TD3) policy gradient reinforcement learning with safety constraints. Our Monte Carlo simulations demonstrate that this co-optimization approach results in modular satellites with better mission performance compared to those designed by evolution-based approaches. Furthermore, this study discusses potential avenues for future research.
}
\maketitle
\thispagestyle{fancy} 

\section*{Nomenclature}
\noindent $\boldsymbol{q}$ \qquad\qquad Quaternion\\
\noindent $\boldsymbol{I}$ \qquad\qquad Moment of inertia\\
\noindent $\boldsymbol{\omega}$ \qquad\qquad Angular velocity\\
\noindent $S$ \qquad\qquad State space\\
\noindent $A$ \qquad\qquad Action space\\
\noindent $P$ \qquad\qquad Transition probability density\\
\noindent $r$ \qquad\qquad Reward fuction\\
\noindent $\gamma$ \qquad\qquad Discounted factor

\section*{Acronyms/Abbreviations}
\noindent MDP \qquad\qquad Markov Decision Process\\
NCA \qquad\qquad Neural Cellular Automata\\
RL \qquad\qquad\quad Reinforcement Learning\\
EA \qquad\qquad\quad Evolutionary Algorithm

\section{Introduction}
As aerospace engineering advances, satellites are increasingly being deployed in complex and dynamic environments \cite{jiao2023advances}. Currently, most satellites are monolithic systems that lack options for maintenance, upgrades, or reconfiguration\cite{kaplan2020modern}. Consequently, its operational life is limited by the lifespan of components or the duration of missions. When a satellite reaches the end of its operational life, it often becomes space debris, which poses risks to other satellites and the International Space Station (ISS). Addressing existing debris is expensive, making the development of satellites that are easier to maintain and reconfigure an urgent priority.

As a promising solution, modular satellite systems can act as a significant paradigm shift \cite{goeller2012modular,kortman2015building,kreisel2019game}. Driven by their adaptability, these systems enable fast customization and alignment with specific mission requirements, revolutionizing satellite deployment and utilization. Recently, a significant trend in modular satellite technology has been the ongoing miniaturization of interchangeable components, driven by advances in electronics and materials science \cite{kopacz2020small}. This development enables the creation of smaller and lighter satellites that are more cost-effective to launch. As computing power becomes increasingly compact and radiation-resistant, sophisticated software can now be deployed in relatively small satellite systems, providing satellites with highly autonomous on-board capabilities \cite{praks2021aalto}. In addition, standardization efforts have also gained momentum, simplifying the development of compatible modules and facilitating the sharing of technology between various research and commercial entities \cite{saeed2020cubesat,kim2021birds,zhang2024cubesat}.

\begin{figure}[t]
    \centering
    \includegraphics[width=\linewidth]{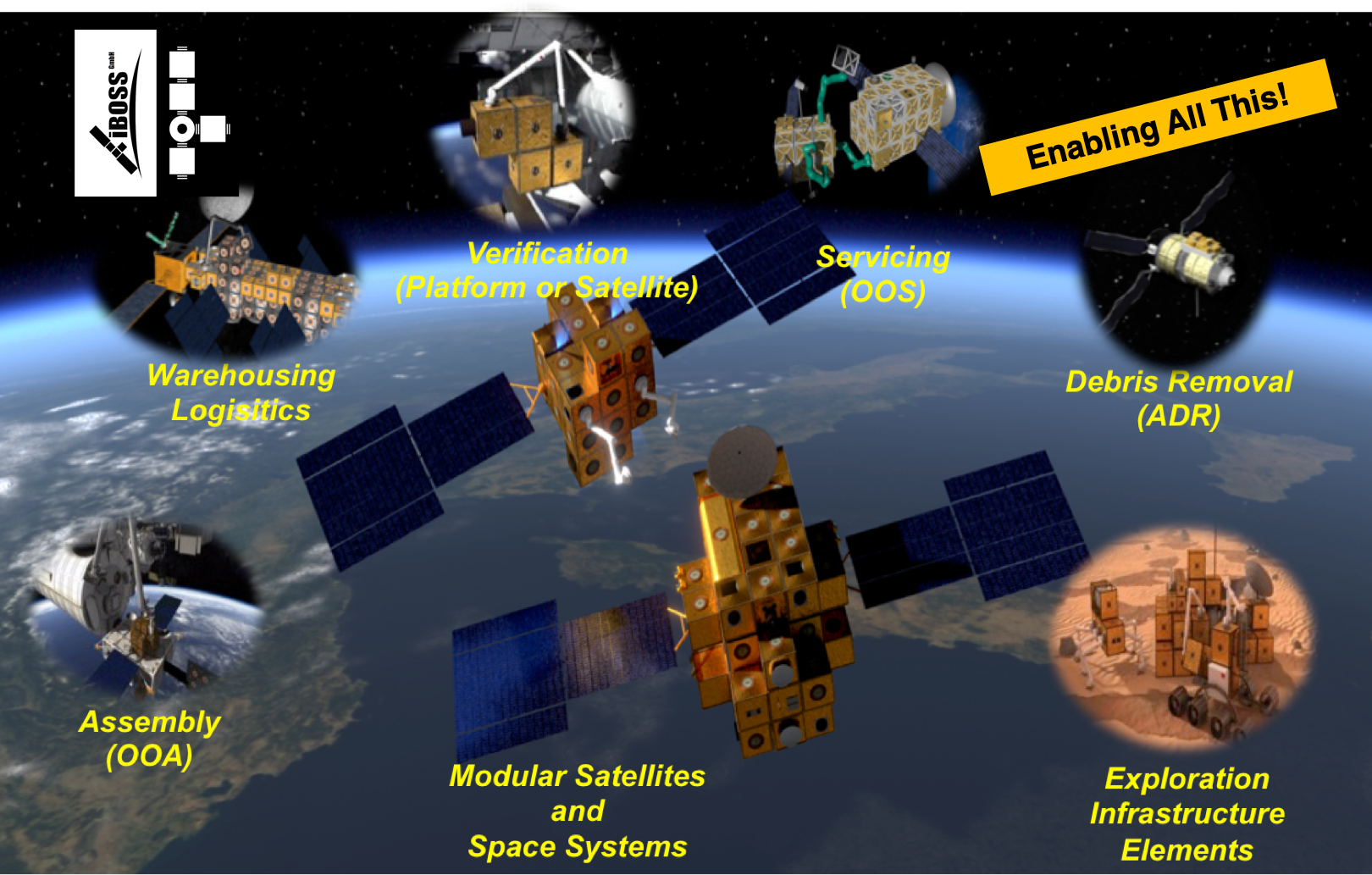}
    \caption{The concept of iBOSS modular satellites \cite{kreisel2019game,zhang2023modularity}, which contains intelligent building blocks for on-orbit satellite servicing.}
    \label{fig1}
\end{figure}

In recent years, several modular satellite projects have been launched, such as Cellsat \cite{tanaka2005autonomous}, ``Intelligent Building-Blocks for On-Orbit Satellite Servicing" (iBOSS) \cite{kosmidis2023metasat}, Phoenix Project \cite{jaeger2014phoenix}, Heterogeneous Cellular Satellite \cite{zhang2023modularity} and OASIS \cite{walker2021orbiting}. Unlike traditional monolithic satellites, as depicted in Fig.\ref{fig1}, modular satellites are composed of individual cubes that each contain different sub-components and perform specific functions. These cubes can be combined to create larger and more complex structures because they are interconnected through multi-function interfaces that facilitate mechanical coupling, energy supply, and communication. Moreover, this inherent flexibility offers a valuable opportunity to optimize the satellite's physical structure (morphology) and its control policy (behavior) simultaneously for critical functions, such as attitude control.

Attitude control \cite{crouch1984spacecraft}, the ability of a satellite to accurately orient itself in space, is vital for a variety of functions, including communication, navigation, and observation. This discipline involves the design and implementation of systems capable of sensing the orientation in a three-dimensional space and making the necessary adjustments to maintain or change this orientation as required by the mission objectives. For several decades, autonomous control of the attitude of the satellite has been a routine practice \cite{elkins2020autonomous,elkins2022bridging}. Beyond the conventional feedback control approach, reinforcement learning (RL) has seen significant advancements in recent years \cite{vedant2019reinforcement,zheng2021reinforcement,tipaldi2022reinforcement}. However, when it comes to modular satellites, attitude control is further complicated by the satellite's evolving configuration. Changes resulting from the addition or reconfiguration of modules, driven by mission-specific requirements, render traditional control strategies less effective.

In summary, addressing the aforementioned issues requires an integrated approach where the morphology of the modular satellite and its control policy are co-optimized for robust and efficient performance. It is crucial to recognize that the development of both physical structures and cognitive capabilities in nature is intricately intertwined \cite{farina2021embodied,cangelosi2022cognitive,wang2023curriculum,wang2023preco}. Designing a modular satellite for a single task is particularly challenging due to: (1) the severe combinatorial explosion within the 3D design space and (2) the existence of incompatible state-action spaces that necessitate training a separate control policy for each satellite morphology. As a result, earlier research \cite{Sims1994Evolving3M, Cheney2013UnshacklingEE,Medvet2021BiodiversityIE,Gupta2021EmbodiedIV} from the Evolutionary Robotics (ER) community frequently tackles these issues by treating the evolution of morphology and control independently, leading to high costs and significant time investment. In contrast to these evolution-based approaches, this study integrates the design and control procedures of modular satellites within a unified Markov Decision Process (MDP) framework, introducing an RL-based co-design approach to explore the extensive design and control spaces, showcasing its effectiveness across different 3D design space settings.

\section{Methods}
\subsection{Reinforcement Learning Algorithm}
Reinforcement learning is a crucial branch of machine learning that focuses on how agents learn the optimal interaction strategy in their environments, which means learning the rules to select the best actions in given states to maximize cumulative rewards. An RL problem can typically be modeled as an MDP, which is a mathematical framework for formulating a discrete-time decision-making process. Generally, an MDP is defined by a five-tuple $<S,A,P,r,\gamma>$. $S$ represents the state space, denoting the set of all possible states of the environment. $A$ represents the action space, indicating the set of all possible actions the agent can take in each state. $P$ is the density of the transition probability, representing the probability of transitioning to state $s_{t+1}$ after taking action $a_t$ in state $s_t$. This probability is typically defined on the basis of the dynamics of the agent and the environment in which it is operating. $r_t$ is the reward function, where $r_t(s_t,a_t)$ denotes the immediate reward received by the agent for taking action $a_t$ in state $s_t$, indicating the desirability of taking a particular action in a given state. $\gamma$ is the discount factor, with a range of $[0,1)$, which specifies the degree to which rewards are discounted over time. Consequently, the agent's goal is to find a policy $\pi_{\phi}(a_t|s_t)$ parameterized by $\phi$ that maximizes the expected cumulative average reward within the time horizon $T$.

\begin{figure}[t]
    \centering
    \includegraphics[width=\linewidth]{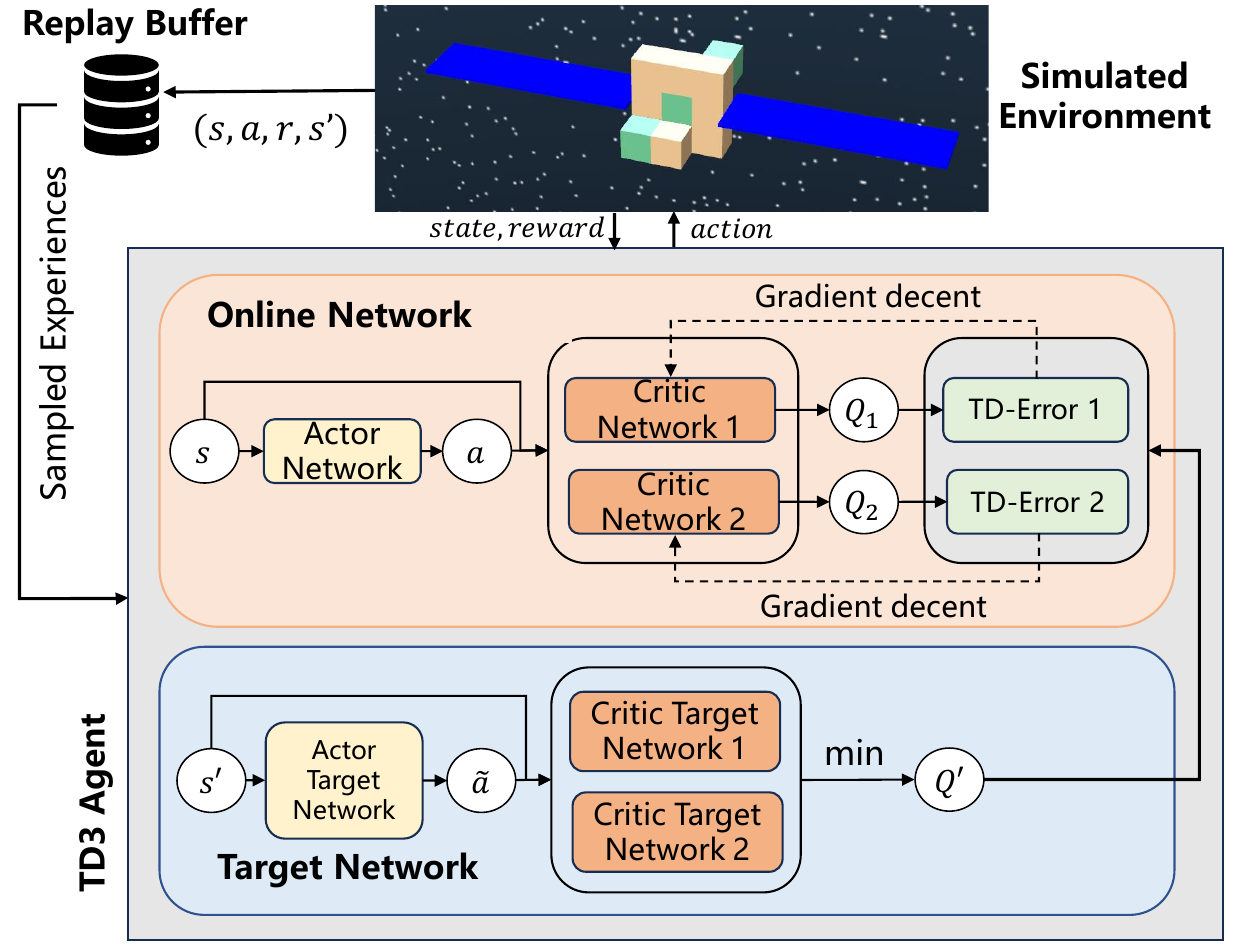}
    \caption{TD3 improves stability in continuous action spaces by using two critic networks, an actor network, and their corresponding target networks. During training, the agent collects experiences by interacting with the environment, storing state transitions in a replay buffer. The training procedure repeats iteratively to refine the policy and improve performance.}
    \label{fig2}
\end{figure}

In this study, we use the Twin Delayed Deep Deterministic Policy Gradient (TD3) algorithm \cite{fujimoto2018addressing,zhang2020model}, a leading-edge technique in reinforcement learning, particularly suited for continuous action spaces, as the optimization strategy for designing and controlling modular satellites. The workflow of the TD3 algorithm is illustrated in Fig.\ref{fig2}. First, all networks and the replay buffer are initialized. Next, in each training iteration, the agent interacts with the environment, where the online network executes a noise-perturbed action chosen by the actor, represented as $a=\pi_\phi(s)+\epsilon,\epsilon\sim N(0,\sigma)$. The agent then observes the resulting reward and state, storing the experience in the replay buffer in the form $(s,a,r,s^{\prime})$. After collecting sufficient experiences, a mini-batch of samples is drawn from the buffer to train the network.

The training process begins with the target network of the actor that generates an action $\Tilde{a}=\pi_{\phi^{\prime}}(s^{\prime})+\epsilon,\epsilon\sim clip(N(0,\sigma),-c,c)$, where $\phi^{\prime}$ represents the parameters of the target network of the actor, and $c$ is the clipping boundary to prevent excessive noise in the target action. The target Q-value is then calculated using the critic target networks, given by the following formula:
\begin{equation}
    y=r+\gamma\underset{i=1,2}{min}Q_{\theta_i^{\prime}}(s^{\prime},\Tilde{a})
\end{equation}
where $\theta^{\prime}_i(i=1,2)$ are the parameters of the two target critic networks. Finally, the parameters of two online critic networks are updated by minimizing the TD error. The optimization objective is performed as:
\begin{equation}
    \theta_i=\underset{\theta_i}{argmin}\frac{1}{N}\sum(y-Q_{\theta_i}(s,a))^2
\end{equation}
where $N$ represents the size of the mini batch. To ensure training stability, the online actor network and all parameters in the target network are updated with a delay, typically every $d$ time step. The objective of the online actor network is to maximize the Q-value estimated by the online critic network. Thus, the actor's parameters are updated using gradient ascent, with the gradient given by:
\begin{equation}
\nabla_\phi J(\phi)=\frac{1}{N}\sum\nabla_aQ_{\theta_1}(s,a)|_{a=\pi_\phi(s)}\nabla_\phi\pi_\phi(s)
\end{equation}
The parameters of the target networks (both actor and critic) are updated with the following formulas:
\begin{equation}
\theta^{\prime}_i=\tau\theta_i+(1-\tau)\theta^{\prime}_i, 
\phi^{\prime}_i=\tau\phi_i+(1-\tau)\phi^{\prime}_i
\end{equation}

In summary, the TD3 algorithm addresses the issue of Q-value overestimation by using two Q-networks. In addition, it adds noise to the target actions to reduce the overfitting of the policy. By employing delayed updates for the networks, TD3 ensures the stability of the Critic's training. These improvements effectively resolve the problems of overestimation of Q-values and policy instability present in DDPG \cite{duan2016benchmarking}, making TD3 a widely used reinforcement learning algorithm for continuous control tasks.

\subsection{Simulation of the Modular Satellites}
\begin{figure}[t]
    \centering
    \includegraphics[width=0.8\columnwidth]{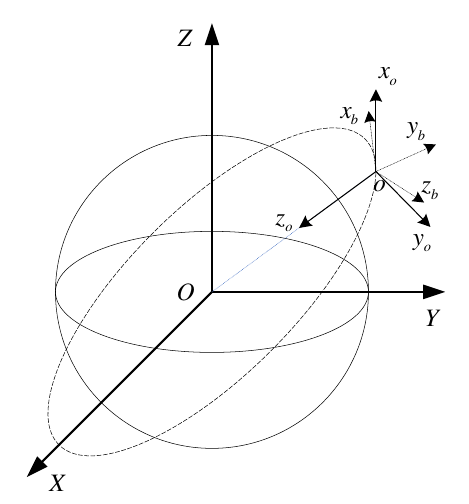}
    \caption{Schematic diagram of coordinate systems used in this work.}
    \label{fig3}
\end{figure}

In this work, we build a simulation environment in a real-world problem setting, where a modular satellite system is modeled similar to the iBoss system \cite{kortman2015building}. To accurately describe the attitude motion of a satellite in orbital space, it is necessary first to establish the geocentric equatorial inertial coordinate system $O-XYZ$, the orbital coordinate system $o-x_oy_oz_o$, and the body-fixed coordinate system $o-x_by_bz_b$. The relative relationships between these three coordinate systems are shown in Fig.\ref{fig3}.
The satellite's attitude is assessed through the relative transformation between its body-fixed coordinate system and the orbital coordinate system. Generally, its attitude can be represented using either Euler angles or quaternions. Euler angles can face singularity problems, whereas quaternions provide more straightforward calculations and avoid these singularities. As a result, we opt to utilize quaternions to describe the attitude. The definition of quaternions is given by the following equation: 
\begin{equation}
\boldsymbol{q}=\left[\begin{array}{l}
q_0 \\
\boldsymbol{q}_{\boldsymbol{v}}
\end{array}\right]=\left[\begin{array}{c}
\cos \frac{\phi}{2} \\
\boldsymbol{e} \sin \frac{\phi}{2}
\end{array}\right]
\end{equation}

In this equation, the real number $q_0$ is the scalar part of the quaternion, and $\boldsymbol{q}_v=\left[\begin{array}{lll}q_x & q_y & q_z\end{array}\right]^T$ is referred to as the vector part. $\boldsymbol{e}=\left[\begin{array}{lll}e_x & e_y & e_z\end{array}\right]^T$ represents the unit vector that describes the axis of rotation, and $\phi$ denotes the rotation angle. By rotating the body-fixed coordinate system counterclockwise around the axis $\boldsymbol{e}$ by the angle $\phi$, it can be aligned with the orbital coordinate system. The kinematic equation of the satellite attitude expressed by quaternions is:
\begin{equation}\label{2}
\dot{q}_0=-\frac{1}{2} \boldsymbol{q}_v^T \boldsymbol{\omega}
\end{equation}
\begin{equation}\label{3}
\dot{\boldsymbol{q}}_v=-\frac{1}{2}\left(\boldsymbol{q}_v^{\times}+q_0 \boldsymbol{E}_{3 \times 3}\right) \boldsymbol{\omega}
\end{equation}
where $\boldsymbol{\omega}=\left[\begin{array}{lll}\omega_1 & \omega_2 & \omega_3\end{array}\right]^T$ represents the projection of the satellite's angular velocity in the body-fixed coordinate system, and $\boldsymbol{E}_{3 \times 3}$ denotes the $3\times3$ identity matrix. $\boldsymbol{q}_v^{\times}$ represents the cross-product matrix of the vector part, which is given by the following expression:
\begin{equation}
\boldsymbol{q}_v^{\times}=\left[\begin{array}{ccc}
0 & -q_z & q_y \\
q_z & 0 & -q_x \\
-q_y & q_x & 0
\end{array}\right]
\end{equation}

For the modular satellite considered here, once its morphology is determined, it can be treated as a single rigid-body satellite. The dynamic equation is expressed as:
\begin{equation}\label{5}
\boldsymbol{I} \dot{\boldsymbol{\omega}}+\boldsymbol{\omega}^{\times} \boldsymbol{I} \boldsymbol{\omega}=\boldsymbol{M}_c+\boldsymbol{M}_d
\end{equation}
where $\boldsymbol{\omega}^{\times}$ represents the cross-product matrix of the angular velocity, calculated in the same way as $\boldsymbol{q}_v^{\times}$. $\boldsymbol{M_c}=\left[\begin{array}{lll}M_{cx} & M_{cy} & M_{cz}\end{array}\right]^T$ represents the total control torque applied by the actuators and $\boldsymbol{M_d}=\left[\begin{array}{lll}M_{dx} & M_{dy} & M_{dz}\end{array}\right]^T$ represents the total disturbance torque acting on the modular satellite. $\boldsymbol{I}$ is its symmetric positive definite inertia matrix, which is expressed as:
\begin{equation}
\boldsymbol{I}=\left[\begin{array}{ccc}
I_{x} & I_{xy} & I_{xz} \\
I_{xy} & I_{y} & I_{yz} \\
I_{xz} & I_{yz} & I_{z}
\end{array}\right]
\end{equation}
where the diagonal elements $I_{x}$, $I_{y}$, and $I_{z}$ represent the inertia moments of the satellite about the three axes of the body-fixed coordinate system. The off-diagonal elements $I_{xy}$, $I_{xz}$, and $I_{yz}$ are referred to as the inertia products. When the body-fixed coordinate system aligns with the principal axes of inertia of the satellite, $I_{xy}=I_{xz}=I_{yz}=0$. At this point, the dynamic equations become significantly simplified. For modular satellites, different module configurations correspond to different inertia matrices. However, as long as the body-fixed coordinate system is aligned with the satellite's principal axes of inertia, the inertia products will be zero, allowing the attitude dynamic equations to still be simplified. 

\begin{figure}[t]
    \centering
    \includegraphics[width=\columnwidth]{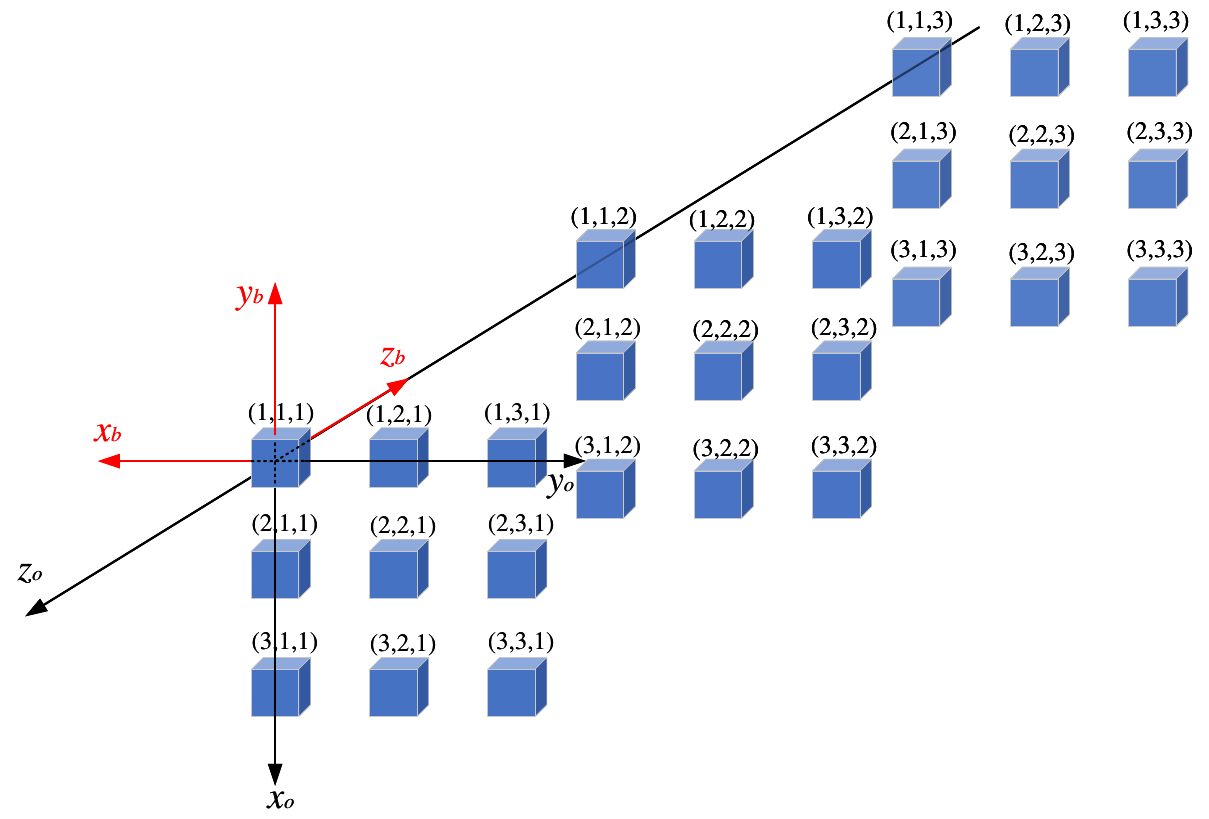}
    \caption{Representation of the modular satellite.}
    \label{fignn}
\end{figure}

We consider the combination of modules shown in Fig.\ref{fignn} and assume that there is no gap between two modules. We establish the reference coordinate system $ox_oy_oz_o$, where the origin $o$ coincides with the center of mass of the module numbered $(1,1,1)$, the $x_o$-axis goes in the same direction as the number of rows increases, the $y_o$-axis and the number of columns increase in the same direction, the $z_o$-axis and the number of layers decrease in the same direction.

We set a single module's mass $m_{sat}=1.0 kg$, its length $l_{sat}=0.1 m$, width $w_{sat}=0.1 m$, and height $h_{sat}=0.1 m$, respectively. The combination of modules is described using a three-dimensional matrix $M_{3d}$ with elements only $0$ and $1$, where $0$ indicates that there are no modules at the corresponding position, $1$ indicates that there exist modules. For the module numbered $(i,j,k)$, the position coordinate of the center of mass is $\mathbf{p}_o^{i,j,k}$ and its coordinate component is $\left[x_o^{i, j, k}, y_o^{i, j, k}, z_o^{i, j, k}\right]^T$, then we have:
\begin{equation}
\mathbf{p}_o^{i,j,k} = \mathbf{M}_{3d}^{i,j,k} \begin{bmatrix}
(i-1) h_{sat} \\
(j-1) l_{sat} \\
-(k-1) w_{sat}
\end{bmatrix}
\end{equation}
Let $n_{sat}$ be the total number of modules, and the position of the module's center of mass with respect to the reference coordinate system is:
\begin{equation}
\mathbf{p}_o^m = \frac{\sum_{i=1}^{3} \sum_{j=1}^{3} \sum_{k=1}^{3} \mathbf{p}_o^{i,j,k}}{n_{sat}} 
\end{equation}

According to the parallel-axis theorem, the moment of inertia $\boldsymbol{I}_o^{i, j, k}$ of the module numbered $(i,j,k)$ relative to each axis of the reference coordinate system of the modular satellite can be calculated by Eq.\ref{eq_i}. Then, the moment of inertia $\boldsymbol{I}^m_o$ of the module relative to each axis of the modular satellite reference coordinate system is calculated by the following formula:
\begin{equation}
\boldsymbol{I}_o^m = \sum_{i=1}^{3} \sum_{j=1}^{3} \sum_{k=1}^{3} \boldsymbol{I}_o^{i,j,k}
\end{equation}

The reference coordinate system is then translated to the centroid of the modular satellite, represented as $o-x_{b t} y_{b t} z_{b t}$, and the coordinate $\boldsymbol{p}_{b t}^{i, j, k}$ of each member module is
\begin{equation}
    \boldsymbol{p}_{b t}^{i, j, k}=\boldsymbol{p}_o^{i, j, k}-\boldsymbol{p}_o^m
\end{equation}

\begin{figure}[ht]
    \centering
    \includegraphics[width=0.8\columnwidth]{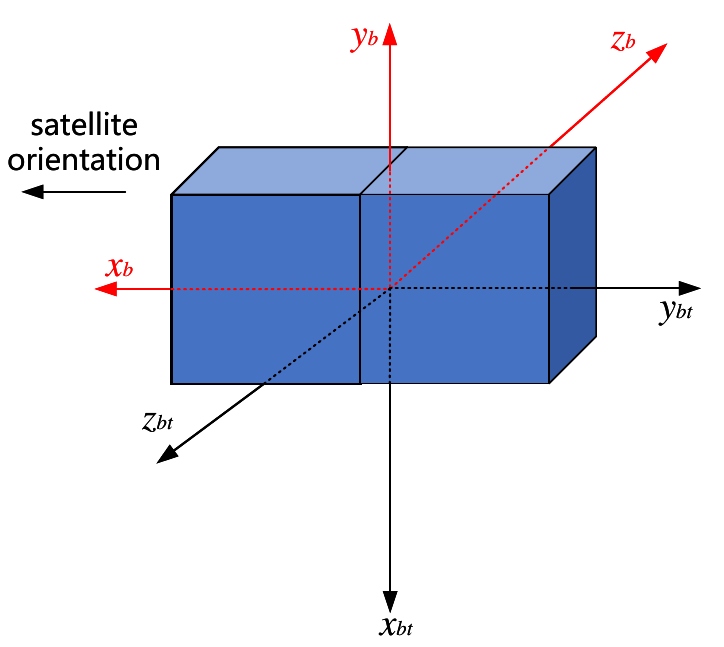}
    \caption{Modular satellite centroid translation reference coordinate system.}
    \label{fig4}
\end{figure}

Similarly, we can calculate the moment of inertia $\boldsymbol{I}_{b t}^{i, j, k}$ of each module using Eq.\ref{eq_ii} and finally the moment of inertia $\boldsymbol{I}_{b}^{i, j, k}$ of each module relative to each axis of the centroid system using Eq.\ref{eq_iii}. The moment of inertia ${I}_{b}^m$ of the modular satellite relative to each axis of the body-fixed coordinate system is:
\begin{equation}
\mathbf{I}_{b}^m = \sum_{i=1}^{3} \sum_{j=1}^{3} \sum_{k=1}^{3} \mathbf{I}_{b}^{i,j,k} \label{14}
\end{equation}
By substituting Eq.\ref{14} into Eq.\ref{5}, the attitude dynamics equation of the modular satellite can be obtained as: 
\begin{equation}
\boldsymbol{I}_b \dot{\boldsymbol{\omega}}+\boldsymbol{\omega}^{\times} \boldsymbol{I}_b \boldsymbol{\omega}=\boldsymbol{M}_c+\boldsymbol{M}_d    
\end{equation}

\subsection{Simulation of the Brain-Body Co-Design Tasks}

\begin{figure*}[t]
    \centering
    \includegraphics[width=\textwidth]{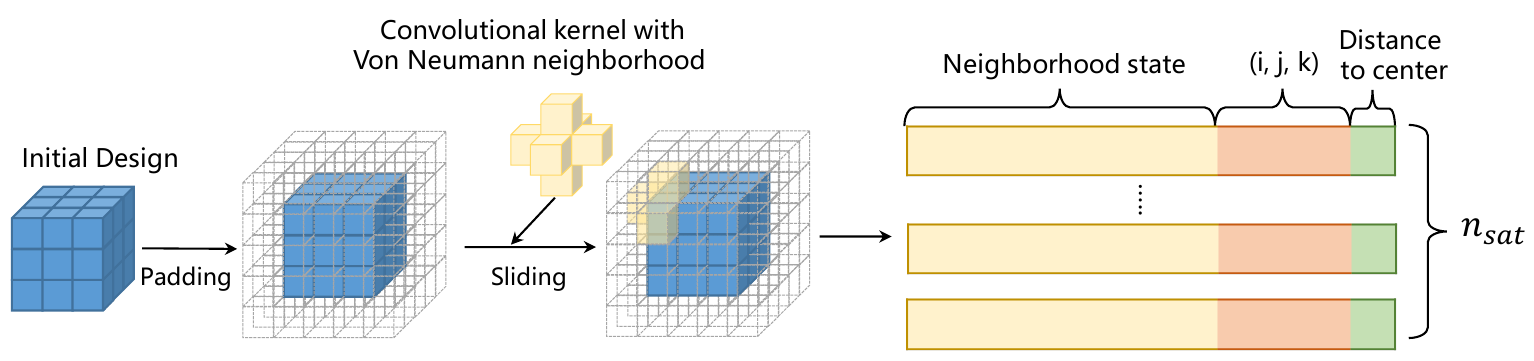}
    \caption{Parameterization of the design space ($3\times3\times3$, for instance). Initially, the design space is surrounded by empty modules. Each module is denoted by a discrete value, reflecting its function. Then, a sliding window is used to get each module's local state, which is composed of its type and the types of its Neumann neighbors. Finally, the design state is formulated as an ordered sequence.}
    \label{fig5}
\end{figure*}

This work combines design and control processes into a single MDP and uses an optimization approach based on RL. At the beginning of each episode, a design policy executes a series of actions to formulate the structure of a modular satellite, without assigning any reward signal to the design policy at this stage. Following this, the produced satellite is consumed by a control policy to gather interaction experiences and environmental rewards, which also feedback as learning signals for the design policy. Through the policy gradient method, both policies can be simultaneously optimized to enhance the performance of the specified attitude control task.

\subsubsection{State Action Spaces Formalization}
To parameterize the design space, we use a Neural Cellular Automata (NCA), which is a MLP that takes in arbitrary modular satellite morphologies and outputs a series of actions to modify them. We focus on modular satellites that include various types of modules. Each module is represented by a discrete value that corresponds to its type (e.g., empty module=$0$, rigid module=$1$, actuator module=$2$). We define the design state $s_{t}^{d}=\{s_{t}^{d_{1}},s_{t}^{d_{2}},...,s_{t}^{d_n}$\} where $s_{t}^{d_{i}}$ for module $i$ is a vector that includes its type and the types of its Neumann neighborhood, together with its position $(i,j,k)$ and distance from the center of mass, as shown in Fig.\ref{fig5}. Following the convention in the literature \cite{vedant2019reinforcement,zheng2021reinforcement,tipaldi2022reinforcement}, the attitude control task is defined as a large-angle slew maneuver that transitions to stabilization in the desired orientation, and we characterize a large-angle slew as targeting an orientation where the angle lies within the range $[30, 150]$ degrees around any rotation axis $\hat{e}$ from the initial orientation.

\begin{figure}[ht]
    \centering
    \includegraphics[width=\columnwidth]{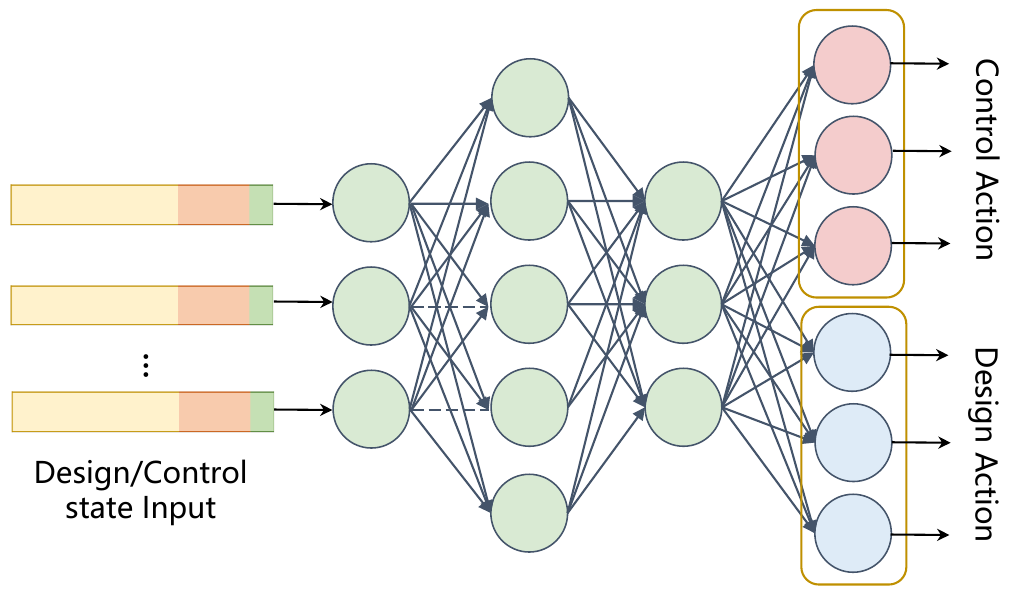}
    \caption{Multi-Layer Perceptron neural network structure with parameter sharing mechanism used in this work.}
    \label{fig6}
\end{figure}

We utilize a reaction-wheel-based attitude control system, where each actuator module corresponds to a reaction wheel. The satellite attitude is represented using Euler parameters (unit quaternions). Thus, the action state vector of the satellite is given as $s_{t}^{c}=\{s_{t}^{c_{1}},s_{t}^{c_{2}},...,s_{t}^{c_n}$\} where $s_{t}^{c_{i}}=\{\boldsymbol{q_e},\dot{\boldsymbol{q_e}},\vec{\omega},position, distance\}$, $\vec{\omega}$ is the angular rotation vector about the body-fixed principal axes. For empty modules, we set $s_{t}^{c_{i}}$ as zero vectors. In practice, $s_{t}^{d_{i}}$ and $s_{t}^{c_{i}}$ are padded with zeros to the same length for convenience in the training process. As illustrated in Fig.\ref{fig6}, our co-design policy is consistently represented as $\pi_{\phi}(a_t|s_t)$, depicted by a neural network with a shared parameter configuration. This indicates that both the NCA and the control policy share the same input and hidden layers, although they differ in their output layers. The vectors for both the design action $a_d$ and the control action $a_c$ have a dimension of $3$. To confine their values within the interval $[-1.0, 1.0]$, we utilize the Tanh activation function on the network output layer. In the design phase, we select the index with the highest value as the candidate module, while in the control phase, each element of the output action vector is limited to a specific peak magnitude by multiplying a scaling factor.

\begin{figure*}[t]
\centering
\includegraphics[width=0.5\textwidth]{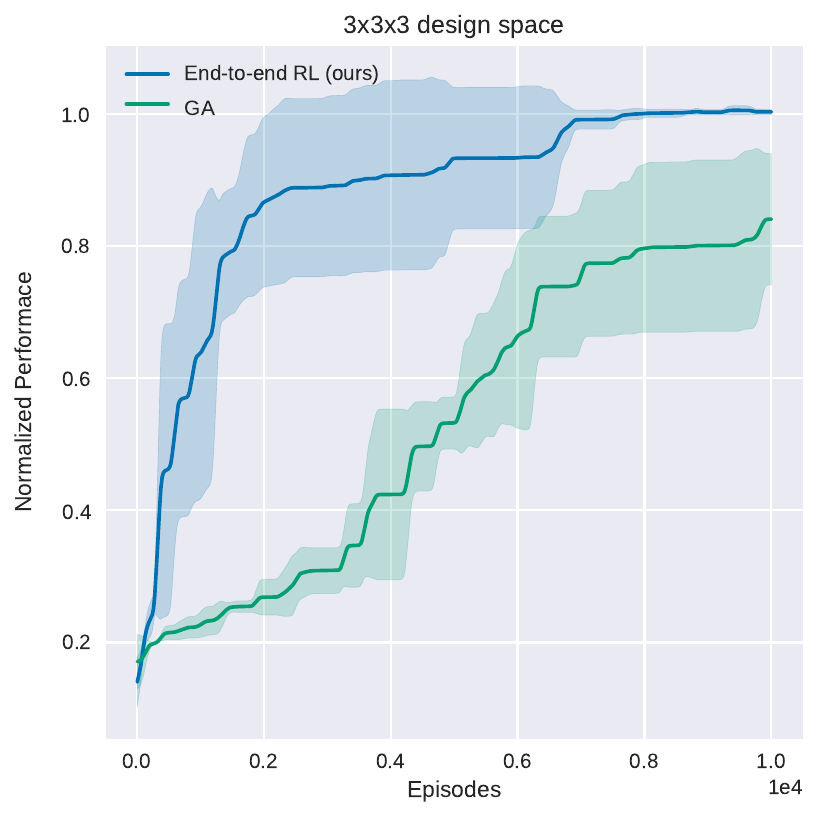}
\includegraphics[width=0.5\textwidth]{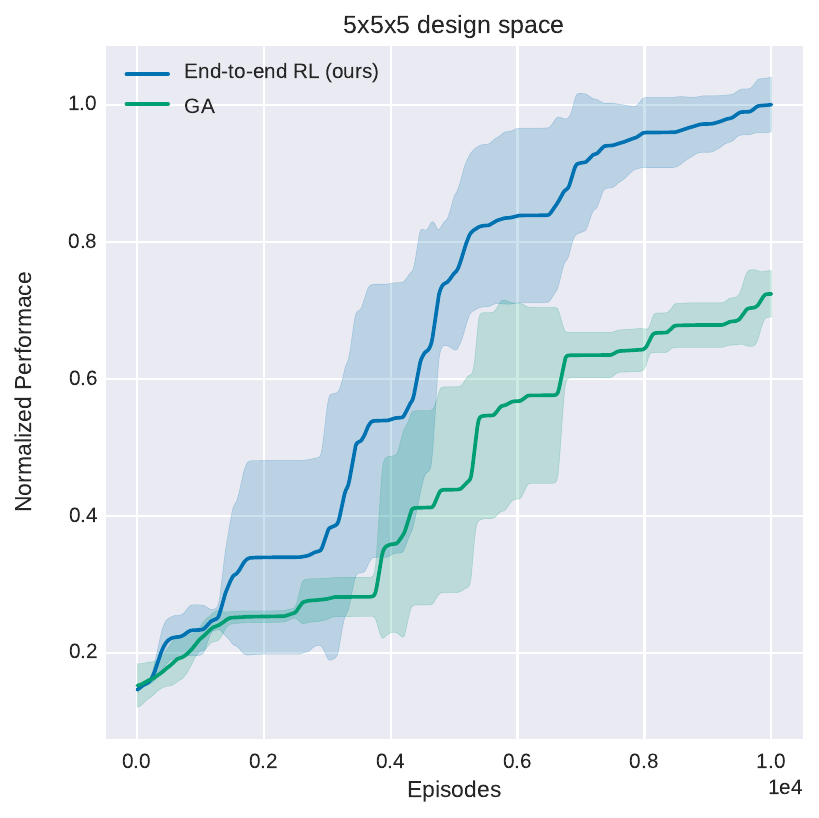} 
\caption{Learning curves of our method and GA in two modular satellite design spaces.}
\label{fitness}
\end{figure*}

\subsubsection{Agent Training Procedure}

The co-design task is packaged according to the standard OpenAI Gym API and is easy to simulate. We limit the maximum episode timestep to $500$ with a ``frame-skipping" interval of $20$. At the beginning of each simulation, a state vector is initialized with random seeds and, at each subsequent time step, the co-design policy neural network determines which action to perform according to the current state vector. The environment simulates this action and returns a new state vector and the corresponding reward. During the design phase, no reward is assigned to the policy, while in the control phase, the produced satellite is initialized at rest, and the following environment reward with safety constraints is given according to the literature \cite{vedant2019reinforcement,zheng2021reinforcement,tipaldi2022reinforcement}.

We train the design policy and the control policy using the TD3 algorithm, which is based on the Actor-Critic architecture. Here, we estimate the value for the entire morphology by averaging the value per module. With the policy gradient method, we use action masks to inform the policy if an element in the output sequence is an actuator module. The hyperparameters of TD3 used in this work are shown in Table \ref{hyperparam}. For training time, it takes around $3$ hours to train our model on a computer with $16$ CPU cores and an NVIDIA RTX $3090$ GPU.

\begin{table}[ht]
    \centering
\begin{tabular}{lr}
\hline 
Hyperparameter & Value \\
\hline Learning rate actor & $3 \mathrm{e}-4$ \\
Learning rate critic & $3 \mathrm{e}-3$ \\
Batch size & $512$ \\
Memory size & $5 \mathrm{e} 5$ \\
Gamma $\gamma$ & $0.99$ \\
Polyak update $\tau$ & $0.995$ \\
Exploration noise $\epsilon$ & 0.2\\
Train frequency & $2$\\
Torque scale for $3\times3\times3$ design space & $0.8$\\
Torque scale for $5\times5\times5$ design space & $1.5$\\
Network Activation Function (hidden) & ReLU\\
Network Activation Function (output) & Tanh\\
Network Hidden Layers (Critic) &[400, 300] \\
Network Hidden Layers (Actor) &[400, 300] \\ 
\hline

\end{tabular}
\caption{Hyperparameters of TD3.}
\label{hyperparam}
\end{table}

\section{Results}
In this section, we demonstrate the effectiveness of our co-design method in attitude control tasks within different modular satellite design spaces ($3\times3\times3, 5\times5\times5$). We seek to answer the following questions: (1) Does our method provide an effective mechanism for learning to design and control modular satellites? (2) How does our method compare with the previous EA-based approach in terms of the performance and the satellite morphologies produced? 

We compare our approach with a popular evolution-based co-design method presented in \cite{Bhatia2021EvolutionGA}. Here, we modify the control optimization using TD3, while the morphology optimization employs a Genetic Algorithm (GA). In this work's setting, GA directly encodes the modular satellite's morphology as a vector where each element is tailored to the module's type in order. It keeps a proportion of satellites with high fitness as survivors at the end of each generation, and then the survivors undergo a mutation conducted by randomly changing each module. We train each proposed method with $6$ different random seeds in two design space settings. During the training process, the average result of $5$ tests of the current co-design policy is reported as the performance of each method. 

\begin{figure}[t]
    \centering
    \includegraphics[width=\columnwidth]{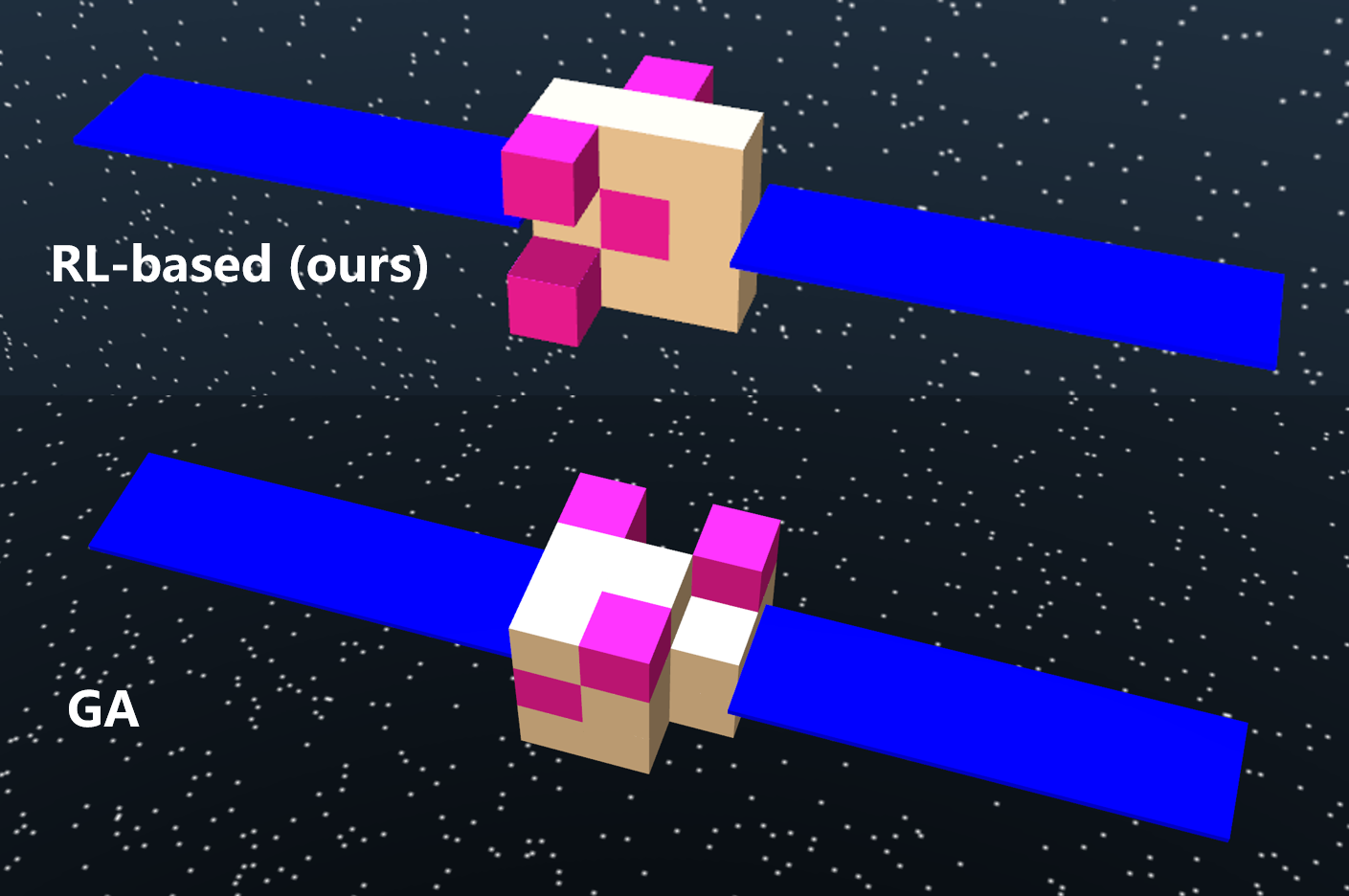}
    \caption{Visualization of modular satellite's morphologies designed by our RL-based method and GA in $3\times3\times3$ design space, where pink modules represent actuators.}
    \label{mor}
\end{figure}

Fig.\ref{fitness} illustrates their learning processes during training. The solid curves represent the mean values and the shaded regions indicate the standard deviations. The figure clearly demonstrates that our proposed co-design method outperforms the baseline in terms of learning speed and final performance. A comparison of the satellite's morphologies designed by each method reveals intriguing distinctions. As shown in Fig.\ref{mor}, GA prefers structures composed of randomly organized modules, leading to much lower task performance, while RL is capable of discovering a satellite body with a symmetry structure, which is essential for maintaining balance in attitude control tasks.

\section{Discussion and Conclusion}
Both the design and control of a satellite are critical to its performance. Despite the well-documented understanding of optimal control by the spacecraft engineering community, less focus is given to identifying the most effective satellite design. Drawing inspiration from biological organisms, where both body structure and brain are vital for task execution, intelligent agents usually need simultaneous optimization of their structure and control mechanisms. By applying this principle to modular satellites, we investigate how designing the satellite's modular structure in conjunction with its control systems can enhance optimal attitude control across different configurations, and demonstrates the effectiveness of our RL-based co-design method on the developed simulation environment, enabling nonexperts to create modular satellites suitable for their requirements. There are many fascinating directions to further advance our proposed technique: we tested our method using a simulator with relatively fundamental modules as a proof-of-concept to show its effectiveness, and there exist many sim-to-real issues when considering real-world settings. For exploring larger design spaces such as $10\times10\times10$, more advanced techniques, for example, transformer, can be an efficient method to dynamically capture the dependencies and relationships between modules, compared to the MLP structure used in this work.

\section*{Acknowledgements}
This work was supported by the National Natural Science Foundation of China No.62293545, the National Natural Science Foundation of China No.62103225, the Shenzhen Natural Science Foundation No.JCYJ20230807111604008, the Guangdong Province Natural Science Foundation No.2024A1515010003, and the Key National Research and Development Program No.2022YFB4701402.

\section*{Appendix A. Calculation of the moment of inertia}
Eq.\ref{eq_i}, Eq.\ref{eq_ii} and Eq.\ref{eq_iii} show the calculation process of the moment of inertia used in this work.

\begin{figure*}[t]
\begin{equation}\label{eq_i}
\boldsymbol{I}_o^{i, j, k}=\boldsymbol{M}_{3 d}^{i, j, k}\left[\begin{array}{l}
\frac{1}{12} m_{\text {sat }}\left(l_{\text {sat }}{ }^2+w_{\text {sat }}{ }^2\right)+m_{\text {sat }}\left(\left(y_o^{i, j, k}\right)^2+\left(z_o^{i, j, k}\right)^2\right) \\
\frac{1}{12} m_{\text {sat }}\left(l_{\text {sat }}{ }^2+w_{\text {sat }}{ }^2\right)+m_{\text {sat }}\left(\left(x_o^{i, j, k}\right)^2+\left(z_o^{i, j, k}\right)^2\right) \\
\frac{1}{12} m_{\text {sat }}\left(l_{\text {sat }}{ }^2+w_{\text {sat }}{ }^2\right)+m_{\text {sat }}\left(\left(x_o^{i, j, k}\right)^2+\left(y_o^{i, j, k}\right)^2\right)
\end{array}\right]
\end{equation} 

\begin{equation}\label{eq_ii}
       \boldsymbol{I}_{b t}^{i, j, k}=\boldsymbol{M}_{3 d}^{i, j, k}\left[\begin{array}{l}
\frac{1}{12} m_{\text {sat }}\left(l_{\text {sat }}{ }^2+w_{\text {sat }}{ }^2\right)+m_{\text {sat }}\left(\left(y_{b t}^{i, j, k}\right)^2+\left(z_{b t}^{i, j, k}\right)^2\right) \\
\frac{1}{12} m_{\text {sat }}\left(l_{\text {sat }}{ }^2+w_{\text {sat }}{ }^2\right)+m_{\text {sat }}\left(\left(x_{b t}^{i, j, k}\right)^2+\left(z_{b t}^{i, j, k}\right)^2\right) \\
\frac{1}{12} m_{\text {sat }}\left(l_{\text {sat }}{ }^2+w_{\text {sat }}{ }^2\right)+m_{\text {sat }}\left(\left(x_{b t}^{i, j, k}\right)^2+\left(y_{b t}^{i, j, k}\right)^2\right)
\end{array}\right] 
    \end{equation}

\begin{equation}\label{eq_iii}
        \boldsymbol{I}_b^{i, j, k}=\boldsymbol{M}_{3 d}^{i, j, k}\left[\begin{array}{l}
\frac{1}{12} m_{\text {sat }}\left(l_{\text {sat }}{ }^2+w_{\text {sat }}{ }^2\right)+m_{\text {sat }}\left(\left(x_{b t}^{i, j, k}\right)^2+\left(z_{b t}^{i, j, k}\right)^2\right) \\
\frac{1}{12} m_{\text {sat }}\left(l_{\text {sat }}{ }^2+w_{\text {sat }}{ }^2\right)+m_{\text {sat }}\left(\left(y_{b t}^{i, j, k}\right)^2+\left(z_{b t}^{i, j, k}\right)^2\right) \\
\frac{1}{12} m_{s a t}\left(l_{\text {sat }}{ }^2+w_{s a t}{ }^2\right)+m_{s a t}\left(\left(x_{b t}^{i, j, k}\right)^2+\left(y_{b t}^{i, j, k}\right)^2\right)
\end{array}\right]
    \end{equation}
\end{figure*}

\bibliography{biblio}

\begin{thebibliography}{10}

\bibitem{jiao2023advances}
Xiaolei Jiao, Jinxiu Zhang, Wenbo Li, Youyi Wang, Wenlai Ma, and Yang Zhao.
\newblock Advances in spacecraft micro-vibration suppression methods.
\newblock {\em Progress in Aerospace Sciences}, 138:100898, 2023.

\bibitem{kaplan2020modern}
Marshall~H Kaplan.
\newblock {\em Modern spacecraft dynamics and control}.
\newblock Courier Dover Publications, 2020.

\bibitem{goeller2012modular}
Michael Goeller, Jan Oberlaender, Klaus Uhl, Arne Roennau, and R{\"u}diger Dillmann.
\newblock Modular robots for on-orbit satellite servicing.
\newblock In {\em 2012 IEEE international conference on robotics and biomimetics (ROBIO)}, pages 2018--2023. IEEE, 2012.

\bibitem{kortman2015building}
M~Kortman, S~Ruhl, Jana Weise, Joerg Kreisel, T~Schervan, Hauke Schmidt, and Athanasios Dafnis.
\newblock Building block based iboss approach: fully modular systems with standard interface to enhance future satellites.
\newblock In {\em 66th International Astronautical Congress (Jerusalem)}, volume~2, pages 1--11, 2015.

\bibitem{kreisel2019game}
Joerg Kreisel, Thomas~A Schervan, Kai-Uwe Schroeder, et~al.
\newblock A game-changing space system interface enabling multiple modular and building block-based architectures for orbital and exploration missions.
\newblock In {\em 70th International Astronautical Congress (IAC)}, 2019.

\bibitem{kopacz2020small}
Joseph~R Kopacz, Roman Herschitz, and Jason Roney.
\newblock Small satellites an overview and assessment.
\newblock {\em Acta Astronautica}, 170:93--105, 2020.

\bibitem{praks2021aalto}
Jaan Praks, M~Rizwan Mughal, R~Vainio, P~Janhunen, J~Envall, P~Oleynik, Antti N{\"a}sil{\"a}, H~Leppinen, P~Niemel{\"a}, A~Slavinskis, et~al.
\newblock Aalto-1, multi-payload cubesat: Design, integration and launch.
\newblock {\em Acta Astronautica}, 187:370--383, 2021.

\bibitem{saeed2020cubesat}
Nasir Saeed, Ahmed Elzanaty, Heba Almorad, Hayssam Dahrouj, Tareq~Y Al-Naffouri, and Mohamed-Slim Alouini.
\newblock Cubesat communications: Recent advances and future challenges.
\newblock {\em IEEE Communications Surveys \& Tutorials}, 22(3):1839--1862, 2020.

\bibitem{kim2021birds}
Sangkyun Kim, Takashi Yamauchi, Hirokazu Masui, and Mengu Cho.
\newblock Birds bus: A standard cubesat bus for an annual educational satellite project.
\newblock {\em JoSS}, 10:1015--1034, 2021.

\bibitem{zhang2024cubesat}
Jiaolong Zhang, Chao Wang, Haoyu Xing, and Jianguo Guo.
\newblock Cubesat standardized modular assembly method and design optimization.
\newblock {\em Acta Astronautica}, 216:370--380, 2024.

\bibitem{tanaka2005autonomous}
Hideyuki Tanaka, Noritaka Yamamoto, Takehisa Yairi, and Kazuo Machida.
\newblock Autonomous assembly of cellular satellite by robot...
\newblock In {\em 56th International Astronautical Congress of the International Astronautical Federation, the International Academy of Astronautics, and the International Institute of Space Law}, pages D1--2, 2005.

\bibitem{kosmidis2023metasat}
Leonidas Kosmidis, Alejandro~J Calder{\'o}n, Aridane~{\'A}lvarez Su{\'a}rez, Stefano Sinisi, Eckart G{\"o}hler, Paco~G{\'o}mez Molinero, Alfred H{\"o}nle, Alvaro~Jover Alvarez, Lorenzo Lazzara, Miguel~Masmano Tello, et~al.
\newblock Metasat: Modular model-based design and testing for applications in satellites.
\newblock In {\em International Conference on Embedded Computer Systems}, pages 347--362. Springer, 2023.

\bibitem{jaeger2014phoenix}
Talbot Jaeger and Walter Mirczak.
\newblock Phoenix and the new satellite paradigm created by hisat.
\newblock In {\em 28th Annual AIAA Conference on Small Satellites}, 2014.

\bibitem{zhang2023modularity}
Zhibin Zhang, LI~Xinhong, LI~Yanyan, HU~Gangxuan, WANG Xun, Guohui Zhang, and TAO Haicheng.
\newblock Modularity, reconfigurability, and autonomy for the future in spacecraft: A review.
\newblock {\em Chinese Journal of Aeronautics}, 36(7):282--315, 2023.

\bibitem{walker2021orbiting}
Christopher~K Walker, Gordon Chin, Susanne Aalto, Carrie~M Anderson, Jonathan~W Arenberg, Cara Battersby, Edwin Bergin, Jenny Bergner, Nicolas Biver, Gordon~L Bjoraker, et~al.
\newblock Orbiting astronomical satellite for investigating stellar systems (oasis): following the water trail from the interstellar medium to oceans.
\newblock In {\em Astronomical Optics: Design, Manufacture, and Test of Space and Ground Systems III}, volume 11820, pages 181--232. SPIE, 2021.

\bibitem{crouch1984spacecraft}
Peter Crouch.
\newblock Spacecraft attitude control and stabilization: Applications of geometric control theory to rigid body models.
\newblock {\em IEEE Transactions on Automatic Control}, 29(4):321--331, 1984.

\bibitem{elkins2020autonomous}
Jacob~G Elkins, Rohan Sood, and Clemens Rumpf.
\newblock Autonomous spacecraft attitude control using deep reinforcement learning.
\newblock In {\em International Astronautical Congress}, 2020.

\bibitem{elkins2022bridging}
Jacob~G Elkins, Rohan Sood, and Clemens Rumpf.
\newblock Bridging reinforcement learning and online learning for spacecraft attitude control.
\newblock {\em Journal of Aerospace Information Systems}, 19(1):62--69, 2022.

\bibitem{vedant2019reinforcement}
James~T Vedant.
\newblock Reinforcement learning for spacecraft attitude control.
\newblock In {\em 70th International Astronautical Congress}, 2019.

\bibitem{zheng2021reinforcement}
Mohong Zheng, Yunhua Wu, and Chaoyong Li.
\newblock Reinforcement learning strategy for spacecraft attitude hyperagile tracking control with uncertainties.
\newblock {\em Aerospace Science and Technology}, 119:107126, 2021.

\bibitem{tipaldi2022reinforcement}
Massimo Tipaldi, Raffaele Iervolino, and Paolo~Roberto Massenio.
\newblock Reinforcement learning in spacecraft control applications: Advances, prospects, and challenges.
\newblock {\em Annual Reviews in Control}, 54:1--23, 2022.

\bibitem{farina2021embodied}
Mirko Farina.
\newblock Embodied cognition: dimensions, domains and applications.
\newblock {\em Adaptive Behavior}, 29(1):73--88, 2021.

\bibitem{cangelosi2022cognitive}
Angelo Cangelosi and Minoru Asada.
\newblock {\em Cognitive robotics}.
\newblock MIT Press, 2022.

\bibitem{wang2023curriculum}
Yuxing Wang, Shuang Wu, Haobo Fu, Qiang Fu, Tiantian Zhang, Yongzhe Chang, and Xueqian Wang.
\newblock Curriculum-based co-design of morphology and control of voxel-based soft robots.
\newblock In {\em The Eleventh International Conference on Learning Representations}, 2023.

\bibitem{wang2023preco}
Yuxing Wang, Shuang Wu, Tiantian Zhang, Yongzhe Chang, Haobo Fu, Qiang Fu, and Xueqian Wang.
\newblock Preco: Enhancing generalization in co-design of modular soft robots via brain-body pre-training.
\newblock In {\em Conference on Robot Learning}, pages 478--498. PMLR, 2023.

\bibitem{Sims1994Evolving3M}
Karl Sims.
\newblock Evolving 3d morphology and behavior by competition.
\newblock {\em Artificial Life}, 1:353--372, 1994.

\bibitem{Cheney2013UnshacklingEE}
Nick Cheney, Robert MacCurdy, Jeff Clune, and Hod Lipson.
\newblock Unshackling evolution: evolving soft robots with multiple materials and a powerful generative encoding.
\newblock In {\em GECCO '13}, 2013.

\bibitem{Medvet2021BiodiversityIE}
Eric Medvet, Alberto Bartoli, Federico Pigozzi, and Marco Rochelli.
\newblock Biodiversity in evolved voxel-based soft robots.
\newblock {\em Proceedings of the Genetic and Evolutionary Computation Conference}, 2021.

\bibitem{Gupta2021EmbodiedIV}
Agrim Gupta, Silvio Savarese, Surya Ganguli, and Li~Fei-Fei.
\newblock Embodied intelligence via learning and evolution.
\newblock {\em Nature Communications}, 12, 2021.

\bibitem{fujimoto2018addressing}
Scott Fujimoto, Herke Hoof, and David Meger.
\newblock Addressing function approximation error in actor-critic methods.
\newblock In {\em International conference on machine learning}, pages 1587--1596. PMLR, 2018.

\bibitem{zhang2020model}
ZhiBin Zhang, XinHong Li, JiPing An, WanXin Man, and GuoHui Zhang.
\newblock Model-free attitude control of spacecraft based on pid-guide td3 algorithm.
\newblock {\em International Journal of Aerospace Engineering}, 2020(1):8874619, 2020.

\bibitem{duan2016benchmarking}
Yan Duan, Xi~Chen, Rein Houthooft, John Schulman, and Pieter Abbeel.
\newblock Benchmarking deep reinforcement learning for continuous control.
\newblock In {\em International conference on machine learning}, pages 1329--1338. PMLR, 2016.

\bibitem{Bhatia2021EvolutionGA}
Jagdeep Bhatia, Holly Jackson, Yunsheng Tian, Jie Xu, and Wojciech Matusik.
\newblock Evolution gym: A large-scale benchmark for evolving soft robots.
\newblock In {\em NeurIPS}, 2021.

\end{thebibliography}

\end{document}